\documentclass[a4paper, 10 pt, conference]{ieeeconf}  

\IEEEoverridecommandlockouts                              

\usepackage{amsmath,amssymb,amsfonts}
\usepackage{algorithmic,algorithm}
\usepackage{xcolor}
\usepackage{url}
\usepackage{siunitx}
\usepackage{graphicx}
\usepackage{subcaption}
\graphicspath{ {./Figures/} }
\usepackage[noadjust]{cite}

\newlength\myindent
\title{\LARGE \bf
BOX3D: Lightweight Camera-LiDAR Fusion for 3D Object Detection and Localization
}

\author{Mario A.V. Saucedo$^{\dagger}$, Nikolaos Stathoulopoulos$^{\dagger}$, Vidya Sumathy, Christoforos Kanellakis \\and George Nikolakopoulos 
\thanks{$\dagger$ These authors contributed equally to this work}
\thanks{The authors are with the Robotics \& AI Group, Department of Computer, Electrical and Space Engineering, Lule\r{a} University of Technology, Lule\r{a} SE-97187, Sweden. 
        {Corresponding author: \tt\small marval@ltu.se}}
\thanks{This work has been partially funded by the research project SP14 ‘Autonomous Drones for Underground Mining Operations’, within the academic programme of the Sustainable Underground Mining (SUM) project, jointly financed by LKAB and the Swedish Energy Agency. Also, is partially funded by the European Union's Horizon Europe Research and Innovation Programme, under the Grant Agreement No. 101091462 m4mining.}
}%

\begin{document}

\maketitle
\thispagestyle{empty}
\pagestyle{empty}

\begin{abstract}

Object detection and global localization play a crucial role in robotics, spanning across a great spectrum of applications from autonomous cars to multi-layered 3D Scene Graphs for semantic scene understanding. This article proposes BOX3D, a novel multi-modal and lightweight scheme for localizing objects of interest by fusing the information from RGB camera and 3D LiDAR. BOX3D is structured around a three-layered architecture, building up from the local perception of the incoming sequential sensor data to the global perception refinement that covers for outliers and the general consistency of each object's observation. More specifically, the first layer handles the low-level fusion of camera and LiDAR data for initial 3D bounding box extraction. The second layer converts each LiDAR's scan 3D bounding boxes to the world coordinate frame and applies a spatial pairing and merging mechanism to maintain the uniqueness of objects observed from different viewpoints. Finally, BOX3D integrates the third layer that supervises the consistency of the results on the global map iteratively, using a point-to-voxel comparison for identifying all points in the global map that belong to the object. Benchmarking results of the proposed novel architecture are showcased in multiple experimental trials on public state-of-the-art large-scale dataset of urban environments.


\end{abstract}

\section{INTRODUCTION}

As technology advances, the demand for accurate and efficient methods to perceive and comprehend three-dimensional scenes becomes increasingly critical. The significance of 3D object detection lies in its pivotal role in autonomous systems, by enhancing the way robots perceive and interact with the three-dimensional world. 
3D object detection enables more sophisticated analysis and understanding of visual data and allows systems to recognize and locate objects in a three-dimensional space, providing a more nuanced and context-aware perception.
The ability to perceive and understand the environment like a 3D space is fundamental for robots to interact seamlessly with their surroundings, whether in outdoor settings or indoors \cite{ma2023,mao2023,wang2023,saucedo2023event,saucedo2023msl3d}.


     \begin{figure*}[!htbp]
        \centering
        \includegraphics[width=\linewidth]
        {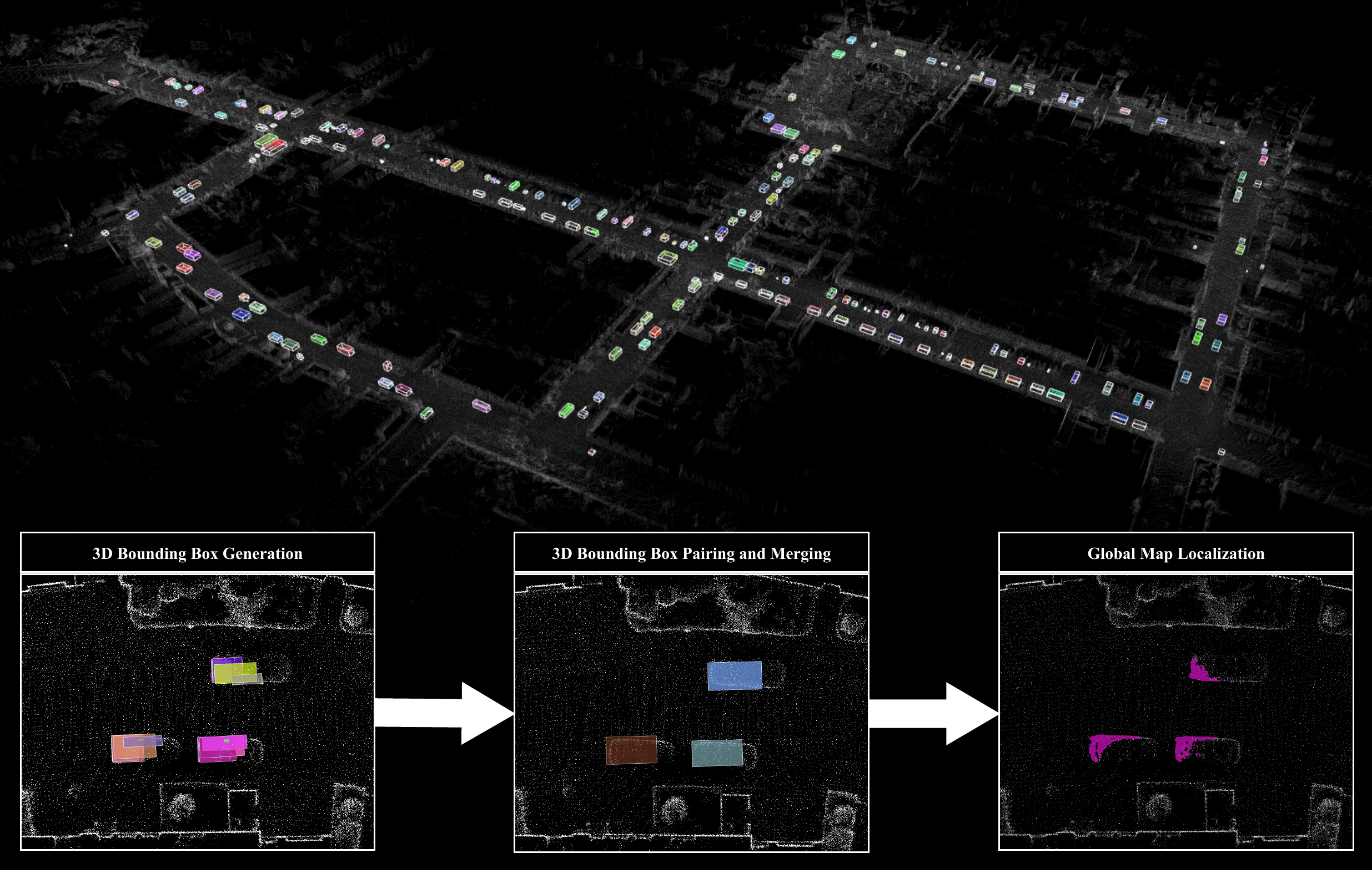}
        \caption{Depiction of the proposed framework BOX3D with the inputs and outputs of each of the layers. Pink points denote the detected objects, while white ones correspond to the rest of the world point cloud.}
        \label{fig:diag}
    \end{figure*}

\subsection*{--- Contributions}

The contributions of this article are summarized as follows. Initially, we present BOX3D a novel framework for camera-LiDAR-based fast object detection and localization, which allows to progressively detect objects in point cloud data. This is achieved due to its three-layer architecture, where the first one optimizes the computation time by generating 2D bounding boxes in the RGB data and then projecting them into the point cloud data to generate 3D bounding boxes. This is done using a state-of-the-art YOLOV8 model for object detection and segmentation in order to generate 2D bounding boxes and segmentation masks in image coordinates. Then the 2D bounding boxes are projected into the LiDAR point cloud using the segmentation masks alongside the intrinsic and extrinsic parameters of the camera. Afterwards, Euclidean clustering is used to filter the points belonging to the background and obtain the set of 3D bounding boxes that identify the bounding volumes of the detected objects with confidence above a specified threshold. 
Once a set of 3D bounding boxes is calculated it is then processed by the second layer, where the 3D bounding boxes of the current LiDAR scan are transformed to world coordinates and then paired with the ones of previous scans based on the degree of overlap. If the percentage of overlap passes a defined threshold these are then merged into a new refined 3D bounding box.
Finally, we use a cluster refining step to segment the points of the world point cloud that correspond to the detected objects. An overview of our approach is presented in Fig.~\ref{fig:diag}.

Secondly, we benchmark the proposed method against a state-of-the-art dataset of large-scale real-life urban environments. Throughout the experimental trials the performance of the framework showcased its merit for fast and precise object detection while building a map of its surroundings. 

\section{RELATED WORK}

In this section, we summarize the current state of the art for PCL-based and image-based 3D detection methods, their main advantages and disadvantages, and how they compare to the proposed fusion. The presented taxonomy divides these algorithms into two main groups, PCL-based, which refers to those algorithms where the detection is done directly over point cloud (PCL) data, and image-based, where the detection occurs on the image provided by a camera and then extrapolates to 3D coordinates with the help of a depth sensor (e.g. a LiDAR or a depth camera).

\subsection{PCL-based Methods}

Voxel-based 3D object detection approaches, as is the case of 
\cite{fan2022}
gridify the irregular point clouds into regular voxels, followed by sparse 3D convolutions to extract high-dimensional features. Despite its effectiveness, voxel-based methods grapple with the trade-off between efficiency and accuracy. Opting for smaller voxels enhances precision but incurs higher computational costs. On the other hand, selecting larger voxels sacrifices potential local details within densely populated voxels.

Other authors have instead opted for point-based 3D object detection approaches 
\cite{yang2020,shi2020,zhou2020,pan2021}, 
that use raw points directly to learn 3D representations. 
Leveraging learning methodologies for point sets mitigates voxelization-induced information loss and capitalizes on the inherent sparsity in point clouds, by confining computations solely to valid data points. However, the imperative for point-based learning operations to be permutation-invariant and adaptive to the input size poses challenges. To achieve this, the model is compelled to learn simple symmetric functions which highly restricts its representation power.

Furthermore, there are point-voxel-based methods 
\cite{shi2020a,he2020,guan2022}, 
which inherently leverage the advantages of fine-grained 3D shape and structural information acquired from points, along with the computational efficiency facilitated by voxels. However, the fusion of point and voxel features typically hinges on voxel-to-point and point-to-voxel transformation mechanisms, incurring non-negligible time costs.


\subsection{Image-based Methods}

Finally, we can find Image-based methods, like in the case of~\cite{zhao2020} where an object detection and identification method that fuses the complementary information obtained by LiDAR and camera is used. Likewise, \cite{pang2020, pang2022} proposed a low-complexity multi-modal fusion framework that operates on the combined outputs of any 2D and any 3D detector previous to Non-Maximum Suppression (NMS). Other authors like~\cite{bai2022} presented a soft-association mechanism to handle inferior image conditions consisting of convolutional backbones and a detection head based on a transformer decoder. 

Furthermore, \cite {li2022} proposed two improvements to classical methods, one to enable accurate geometric alignment between LiDAR points and image pixels, and another that leverages cross-attention to dynamically capture the correlations between image and LiDAR features during fusion. On the other hand,~\cite{qin2023} proposed a novel training strategy that provides an auxiliary feature level supervision for effective LiDAR and camera fusion and significantly boosts detection performance.
In contrast to the aforementioned works, our framework does not rely on multiple networks for the fusion process, instead, leverages different techniques to map 2D bounding boxes to 3D coordinates while progressively refining the generated 3D bounding boxes on the global map.
 
    \begin{figure*}[!htbp]
        \centering
        \includegraphics[width=1.0\textwidth]{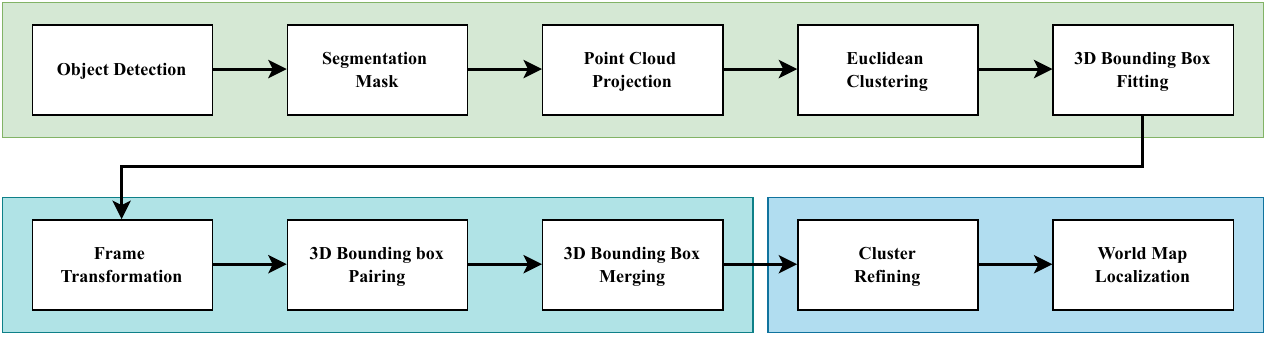}
        \caption{Functional block diagram of the proposed framework for lightweight object detection and localization based on camera-LiDAR fusion.}
        \label{fig:graph}
    \end{figure*}

\section{METHODOLOGY}

An overview of the proposed framework is presented in the functional block diagram in Fig. \ref{fig:graph}. The framework is composed of three layers, the first one generates bounding boxes and segmentation masks in image coordinates and transforms them into LiDAR coordinates to generate 3D bounding boxes while keeping the computational time low, meanwhile, the second one evaluates the degree of overlap between the 3D bounding boxes of the current detection and the previously detected ones, if the degree of overlap surpass a threshold the corresponding 3D bounding boxes are fused to generate a new refined 3D bounding box, finally, the last layer evaluates the points of the world point cloud that correspond to each detected object and the location of the objects in world coordinates. The rest of this section explains in detail all the sub-components of the proposed framework.

\subsection{Coordinate Frames and Transformations}

The world frame $\mathbb{W}$ is fixed and defines the workspace of the robotic platform, the LiDAR frame $\mathbb{L}$ is located on the LiDAR sensor, while the image frame $\mathbb{C}$ is attached on the camera sensor. The projection of a point $p^\mathbb{L} = (x, y, z)$ to $\mathbb{C}$ is performed using the homogeneous transformation matrix:
\begin{equation}
    \begin{bmatrix}
     u\\
     v\\
    1
    \end{bmatrix}
    =
    \underbrace{
     \begin{bmatrix}
     f_x&  0&  c_x\\
     0&  f_y&  c_y\\
     0&  0&  1\\
    \end{bmatrix}
    }_K
    \underbrace{
    \begin{bmatrix}
     r_{11}&  r_{12}& r_{13}& t_{x}\\
     r_{21}&  r_{22}& r_{23}& t_{y}\\
     r_{31}&  r_{32}& r_{33}& t_{z}\\
    \end{bmatrix}
    }_{[R|t]}
    \begin{bmatrix}
     x\\
     y\\
     z\\
    1
    \end{bmatrix}
    \label{eq:proj}
\end{equation}

where the vector $[x, y, z,1]^T$ represents the 3D coordinates of any given point on the point cloud, $R$ and $t$ are the rotation matrix and the translation vector, respectively, between the camera and the LiDAR sensor, $f_x$ and $f_y$ are the focal lengths of the camera with principal point $(c_x,c_y)$, and $p^\mathbb{C} = (u,v)$ is the projection of the point $p^\mathbb{L}$ to 2D image coordinates.

Likewise, the transformation of a point $p^\mathbb{L}$ to $\mathbb{W}$ is performed using the homogeneous transformation matrix $T^{\mathbb{W}}_{\mathbb{L}}$. 
In this work, we utilize Direct Lidar Odometry (DLO) \cite{chen2022direct} fused with Inertial Measurement Unit (IMU) data to estimate the transformation matrix $T^{\mathbb{W}}_{\mathbb{L}}$. 
For the notations in the rest of the article the superscript $(\cdot)^{\#}$ denotes the reference frame. 
 
     \begin{figure*}[!htbp]
        \centering
             \begin{subfigure}[b]{0.49\textwidth}
                 \centering
                 \includegraphics[width=\textwidth]{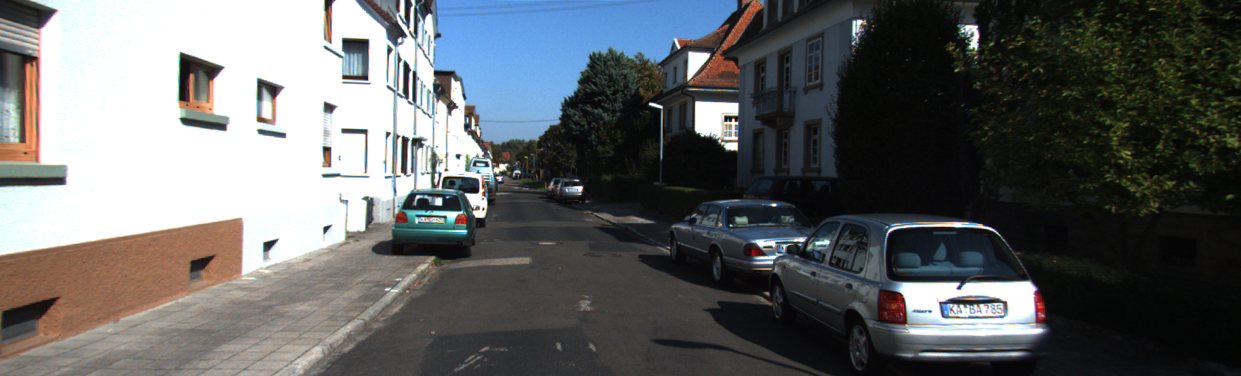}
                 \caption{Input image}
                 \label{fig:rgb}
             \end{subfigure}
             \hfill
             \begin{subfigure}[b]{0.49\textwidth}
                 \centering
                 \includegraphics[width=\textwidth]{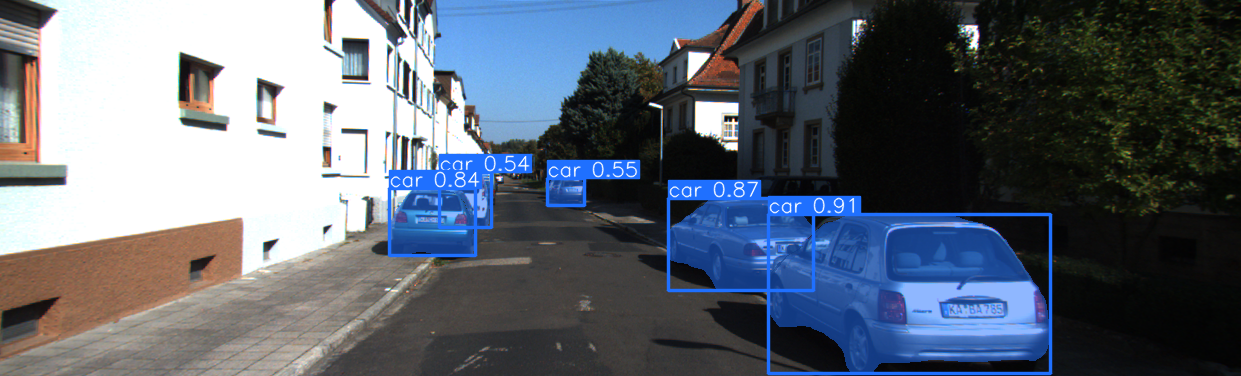}
                 \caption{Object detection bounding boxes}
                 \label{fig:yolo}
             \end{subfigure} 
             \par\bigskip
        \vspace{-0.8em}
             \begin{subfigure}[b]{0.49\textwidth}
                 \centering
                 \includegraphics[width=\textwidth]{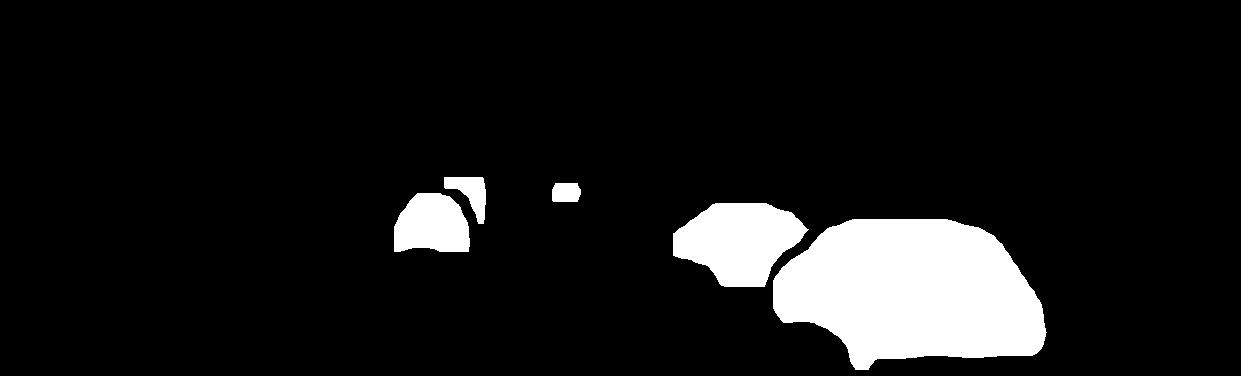}
                 \caption{Segmentation mask}
                 \label{fig:mask}
             \end{subfigure}
             \hfill             
             \begin{subfigure}[b]{0.49\textwidth}
                 \centering
                 \includegraphics[width=\textwidth]{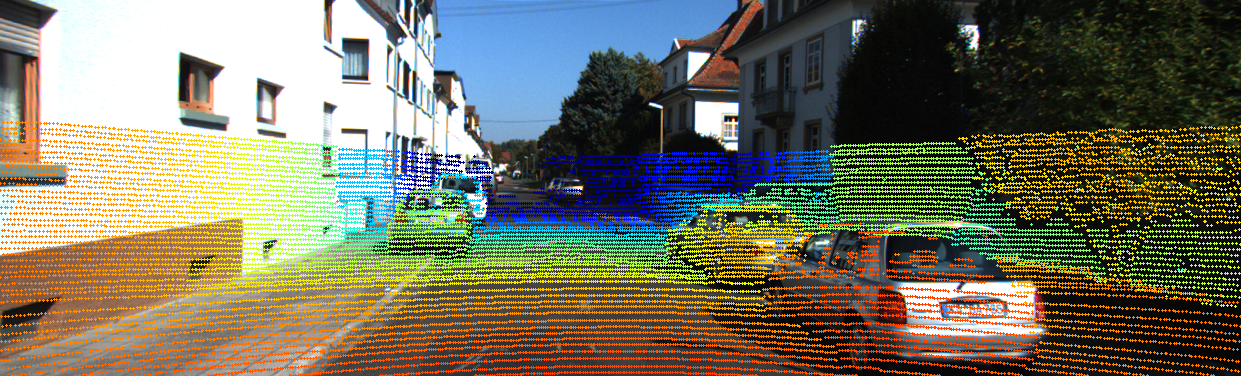}
                 \caption{Point cloud overlay}
                 \label{fig:overlay}
             \end{subfigure}
        \caption{Example of the input (a) and outputs of each step on the 3D bounding box generation module, where 2D bounding boxes (b) are mapped to 3D coordinates using the segmentation mask (c) to label the points on the projected point cloud~(d).}
        \label{fig:img}
    \end{figure*}


\subsection{3D Bounding Box Generation}
\label{sec:3db}

The proposed framework first uses a state-of-the-art YOLOV8n model trained on the COCO dataset~\cite{cocodataset} for detection and segmentation. The model takes as input an RGB image and outputs a set of $m \in \mathbb{Z^+}$ bounding boxes where each box is described by a vector consisting of the $x$ and $y$ coordinates of the box's center, its width and height, 80 class confidences, and 32 mask weights, in addition to 32 prototype masks of $160\times160$ pixels for each of the boxes.

Then we apply Non-Maximum Suppression (NMS) on the box's class confidence to get the subset of boxes $\mathcal{B}^{\mathbb{C}} = \{b^{\mathbb{C}}_1,..., b^{\mathbb{C}}_n \,|\, n \in \mathbb{Z^+}\}$ over a specified confidence threshold (Fig. \ref{fig:yolo}). 
Additionally, we obtain the final segmentation mask of the resulting boxes by multiplying each prototype mask with its corresponding mask weight and then summing all these products together. The resulting mask is then resized to the input frame dimensions and eroded with a morphological filter (Fig. \ref{fig:mask}). The erosion process aims to minimize the amount of false positives on the final segmentation mask. 

Subsequently, the respective LiDAR scan (i.e., the one synchronized with the corresponding frame) is projected into image coordinates using Equation~\ref{eq:proj} (Fig. \ref{fig:overlay}). Each projected point is then assigned an instance label if it lies within a segmented mask, otherwise is categorized as background.
Afterwards, we apply Euclidean clustering
, to each of the resulting instance labeled point clusters excluding background. This is done to filter out any outliers from the projection.  
For the specifics of our framework, we assume that the cluster 
with the greater amount of points is the one with the points belonging to the object. 
This process provides the set of labeled point clusters $\mathcal{D}^{\mathbb{L}} = \{d^{\mathbb{L}}_1,..., d^{\mathbb{L}}_n \,|\, n \in \mathbb{Z^+}\}$ for each of the detected objects. Furthermore, these clusters are then used to fit the set of 3D bounding boxes $\mathcal{B}^{\mathbb{L}} = \{b^{\mathbb{L}}_1,..., b^{\mathbb{L}}_n \,|\, n \in \mathbb{Z^+}\}$.

\subsection{3D Bounding Box Pairing and Merging}


The previous steps are repeated for every frame and LiDAR scan pair, but due to the moving nature of the robotic platform, at every time instance $t$ different objects may be visible from the Field Of View (FOV) of the camera or within the range of the LiDAR, meaning that the set of bounding boxes  $\mathcal{B}^{\mathbb{C}}_t$ will dynamically change, and by consequence the set of clusters $\mathcal{D}^{\mathbb{L}}_t$ and the subsequent set of 3D bounding boxes $\mathcal{B}^{\mathbb{L}}_t$ will also change. 
In order to avoid generating multiple 3D bounding boxes for the same instance of a detected object and to increase the accuracy of the framework, the set of 3D bounding boxes $\mathcal{B}^{\mathbb{L}}_t$ obtained at each time instance $t$ is merged with the set of 3D bounding boxes of previously detected objects to generate a single set $\mathcal{B}^{\mathbb{W}}$ containing the unique 3D bounding box of every object.

The first step is to transform the sets $\mathcal{D}^{\mathbb{L}}_t$ and $\mathcal{B}^{\mathbb{L}}_t$ from the LiDAR frame of reference $\mathbb{L}$ to the world frame of reference $\mathbb{W}$ with the homogeneous transformation matrix $T^{\mathbb{W}}_{\mathbb{L}}$ provided by the DLO framework \cite{chen2022direct}.
Nevertheless, the set of clusters $\mathcal{D}^{\mathbb{W}}_t$ and $\mathcal{D}^{\mathbb{W}}_{\tau < t}$ are unlikely to intercept even in successive scans, despite the likeliness for clusters pairs between $\mathcal{D}^{\mathbb{W}}_t$ and $\mathcal{D}^{\mathbb{W}}_{\tau < t}$ to correspond to the same instance of a detected object, on the other hand, the corresponding pair of 3D bounding boxes on $\mathcal{B}^{\mathbb{W}}_t$ and $\mathcal{B}^{\mathbb{W}}_{\tau < t}$ are likely to present an overlap still. 
Based on the aforementioned, when the percentage of overlap of any given pair $b^{\mathbb{W}}_t \cap b^{\mathbb{W}}_{\tau < t}$ surpass a defined threshold the two 3D bounding boxes are assumed to belong to the same object instance and are merged. The merging is done by fitting a new 3D bounding box that encompasses all the points of the two corresponding clusters on $\mathcal{D}^{\mathbb{W}}_t$ and $\mathcal{D}^{\mathbb{W}}_{\tau < t}$ 
This gives as an output the refined set of 3D bounding boxes $\mathcal{B}^{\mathbb{W}}$. 


\subsection{Global Map Localization}
\label{sec:gml}

Extracting the precise location of any given object in the global map requires the knowledge of all the points of the world point cloud that belong to the same, however, as the robot moves the number of points that constitute the world point cloud increases. This means that in frames where YOLOV8n fails to detect an object there is still a chance that points belonging to that undetected object will be registered on the world point cloud. Furthermore, objects outside the FOV of the camera may still fall within the sensor range of the LiDAR, thus even detected objects may have points added at future updates of the world point cloud. 
These points will not be part of any cluster on the sets $\mathcal{D}^{\mathbb{W}}_t$ and $\mathcal{D}^{\mathbb{W}}_{\tau < t}$ which means that the union of the points belonging to both clusters in any given pair of matched clusters $d^{\mathbb{W}}_t \cup d^{\mathbb{W}}_{\tau < t}$ will fail to encompass all the points on the world point cloud that belong to the respective object instance.
 
To overcome this we do a cluster refining step after each 3D bounding box merging step, in which 
we obtain the set of unique clusters $\mathcal{D}^{\mathbb{W}}$ for every detected object. This is done by encompassing the points of the world point cloud that fall within the voxel region of side $r$ around each of the points of every pair of matched clusters $d^{\mathbb{W}}_t \cup d^{\mathbb{W}}_{\tau < t}$~\cite{stathoulopoulos2023irregular} 
The resulting set $\mathcal{D}^{\mathbb{W}}$ is then used to determine the location of every detected object on the global map by estimating the centroid of each cluster.


\section{RESULTS}

\subsection{Dataset and Benchmark}

\begin{figure*}[!h]
    \centering
    \includegraphics[width=1.0\textwidth]{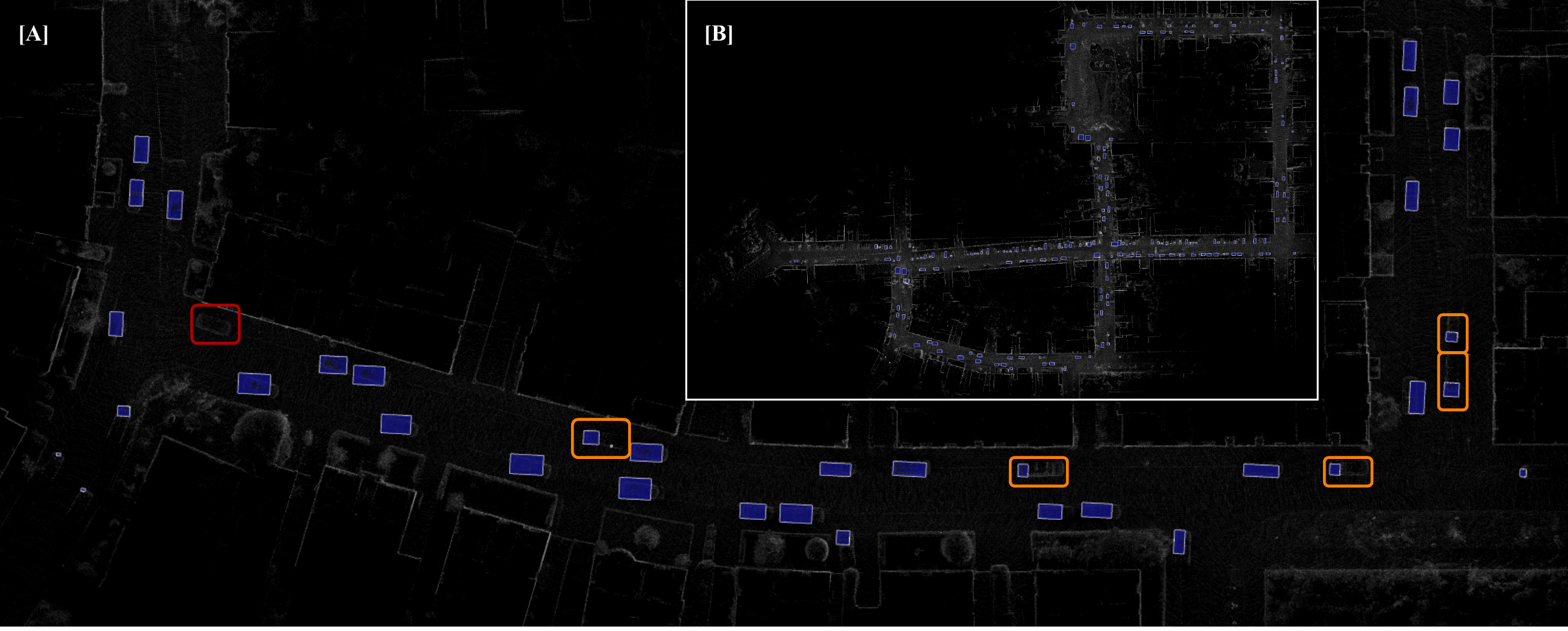}
    \caption{Visualization of the 3D bounding boxes of the detected objects on the global map [B] and of different instances of \textcolor[HTML]{FF0000}{miss-detection} and \textcolor[HTML]{FF8000}{partial detection} [A].}
    \label{fig:res1}
\end{figure*}

\begin{figure*}[!h]
    \centering
    \includegraphics[width=1.0\textwidth]{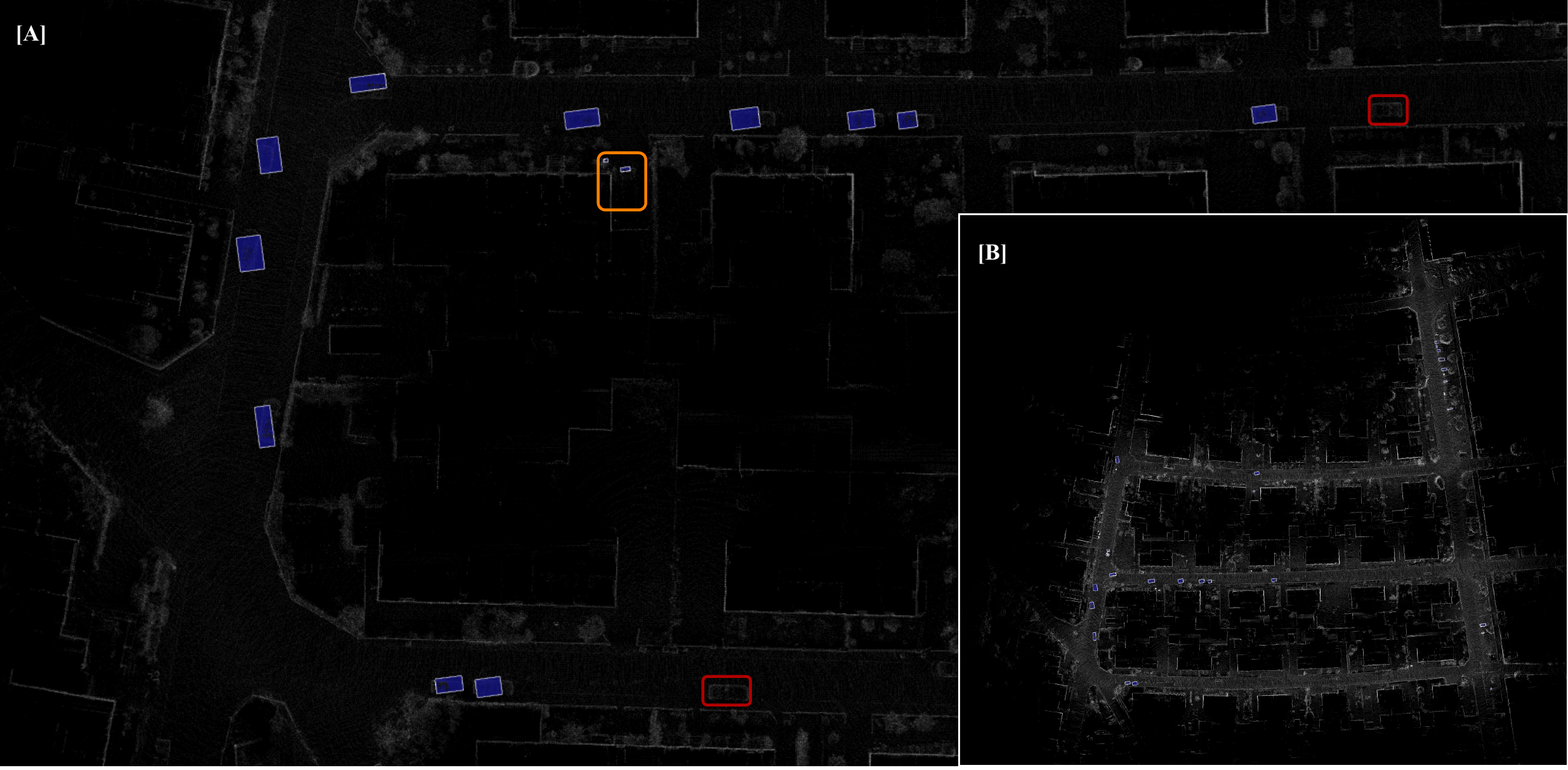}
    \caption{Visualization of the 3D bounding boxes of the detected objects on the global map [B] and of different instances of \textbf{\textcolor[HTML]{FF0000}{miss-detection}} and \textbf{\textcolor[HTML]{FF8000}{partial detection}} [A].}
    \label{fig:res2}
\end{figure*}

We evaluated the performance of our framework using the KITTI dataset~\cite{Geiger2012CVPR}, which includes images and LiDAR scans of large-scale real-life urban environments, alongside ground-truth 3D bounding boxes. The evaluation metrics for this work were the average computation time of each layer and the mean Intersection over Union (mIoU), measured on a 12th Gen Intel(R) Core(TM) i9-12900KF. For the time comparison we used the YOLOv8n model, the results are presented in Table \ref{tab1}, where we can see that the second layer is the most computationally heavy, followed by the first layer. Due to the fact that the algorithm in the second layer needs to check all previously generated 3D bounding boxes to make the merging, meaning that the computation time of this layer increases with every new detected object. 

\begin{table}[!htbp]
\caption{Time Comparison}
        \vspace{-1em}
\begin{center}
    \begin{tabular}{|c|c c c|}
    \hline
    \textbf{Layer}& \textbf{I}& \textbf{II}& \textbf{III} \\
    \hline
    \hline
    \textbf{Time} & $2.5ms$& $8.4ms$& $0.048ms$\\
    \hline
    \end{tabular}
\label{tab1}
        \vspace{-1.5em}
\end{center}
\end{table}

In addition, Table \ref{tab2} presents the obtained mIoU of the proposed framework using different YOLOv8 model sizes, the mIoU of the YOLO model is also listed for comparison. In general, the proposed framework presents close values of mIoU to the ones of the selected object detection model. This shows, as expected, that the performance of the framework increases according to the performance of the model.

\begin{table}[!htbp]
\caption{Performance Evaluation}
        \vspace{-1em}
\begin{center}
    \begin{tabular}{|c|c c c|}
    \hline
    \textbf{Model}& \textbf{nano}& \textbf{small}& \textbf{medium} \\
    \hline
    \hline
    YOLOv8 & $37.3$& $44.9$& $50.2$\\
    BOX3D & $29.7$& $38.6$& $43.2$\\
    \hline
    \end{tabular}
\label{tab2}
        \vspace{-1.5em}
\end{center}
\end{table}

Finally, Fig. \ref{fig:res1} and Fig. \ref{fig:res2} show visualizations of the detected 3D bounding boxes with the proposed method using YOLOv8n as the object detection model. In the figures, the red boxes indicate miss-detection, i.e. undetected objects, while the orange ones indicate partial detection, i.e. bounding boxes that enclose less than 50\% of the object.


\addtolength{\textheight}{-1cm} 

\subsection{Limitations}


The main limitations of this framework stem from some shortcomings on the fusion that originate from the individual sensors described below:

    \textbf{Segmentation Noise}. The process of semantic segmentation is known to incorrectly mark pixels close to the borders of an object as part of the object. The erosion process described in Section~\ref{sec:gml} helps to mitigate the effects of these defects. However, through the erosion it might happen that points belonging to an object may be marked as background during the projection of the point cloud over the segmentation mask.
    
    \textbf{Field Of View}. One of the advantages of using a LiDAR sensor is its $360^{\circ}$ FOV nature, while other sensors like cameras will usually have substantially smaller FOV. Our framework is unable of taking full advantage of this property during the first layer, while the refinement step in Section~\ref{sec:gml} takes into consideration the points from the LiDAR point cloud that are outside of the camera's FOV. Nevertheless, it might still miss small subsets of points that belong to the objects.


\section{CONCLUSIONS}

This article proposed the BOX3D framework for lightweight object detection and localization based on camera-LiDAR fusion. BOX3D is a novel three-layered architecture, where the first layer focuses on the computation efficiency of 3D bounding box generation from 2D object segmentation and LiDAR point cloud projection, the second layer unifies subsequent 3D bounding boxes into a unique bounding box for each object instance through a spatial merging and pairing, and finally the third layer utilizes the complete LiDAR-based global point cloud information to fine-tune each unique 3D bounding box cluster with all neighboring points. Furthermore, the proposed scheme was experimentally tested in large-scale real-world urban environments provided by the KITTI dataset.The test results proves the competence of the proposed BOX3D architecture in the object detection and
localization tasks.






\bibliographystyle{ieeetr}
\bibliography{References}

\end{document}